# Finding Words Associated with DIF: Predicting Differential Item Functioning using LLMs and Explainable AI

Hotaka Maeda[1] and Yikai Lu[2]

[1]Smarter Balanced, University of California-Santa Cruz

[2]Department of Psychology, University of Notre Dame

We fine-tuned and compared several encoder-based Transformer large language models (LLM) to predict differential item functioning (DIF) from the item text. We then applied explainable artificial intelligence (XAI) methods to these models to identify specific words associated with DIF. The data included 42,180 items designed for English language arts and mathematics summative state assessments among students in grades 3 to 11. Prediction $R^2$ ranged from .04 to .32 among eight focal and reference group pairs. Our findings suggest that many words associated with DIF reflect minor sub-domains included in the test blueprint by design, rather than construct-irrelevant item content that should be removed from assessments. This may explain why qualitative reviews of DIF items often yield confusing or inconclusive results. Our approach can be used to screen words associated with DIF during the item-writing process for immediate revision, or help review traditional DIF analysis results by highlighting key words in the text. Extensions of this research can enhance the fairness of assessment programs, especially those that lack resources to build high-quality items, and among smaller subpopulations where we do not have sufficient sample sizes for traditional DIF analyses.

## Introduction

Fair assessments are paramount in education. Students should not be given any advantage or disadvantage based on their demographic backgrounds. Items that exhibit such bias can be identified using differential item functioning (DIF) analysis. DIF is when correct item response probability depends on the examinee's demographic background, even after accounting for ability (Holland & Wainer, 2012). Although DIF analysis has become a standard procedure for evaluating item fairness in educational and psychological assessment (AERA et al., 2014), it is difficult to perform for several reasons. First, it requires data from at least hundreds, if not thousands of examinees for each group for each item. Obtaining the sample size is especially challenging for minority groups that make up a smaller proportion of the population. Second, identified DIF items are not always considered unfair, and should not be automatically removed from assessments. For example, DIF can simply indicate legitimate differences in ability between groups due to sub-domains present in the test blueprint by design (Osterlind & Everson, 2009). Evaluating DIF items require thoughtful qualitative human review (AERA et al., 2014), but the source of DIF is often difficult to identify (Angoff, 1993).

Recently, large language models (LLMs) based on Transformers (Vaswani et al., 2017) have quickly grown in their capability to understand text data (Devlin et al., 2019). LLMs are commonly recognized as *black-box* machine learning (ML) models due to the complexity of their model architecture. However, explainable artificial intelligence (XAI) methods have been developed to provide interpretable explanations to better understand predictions or decisions made by these black-box models (Kokhlikyan et al., 2020). More specifically, XAI methods can identify and attribute the influence of specific features of an ML model (e.g., variables or tokens) on its predictions. If LLMs can be used to predict DIF, and XAI methods can describe how the model makes such prediction, then we can identify specific words in the items that are associated with DIF (see Figure 1). As a consequence, the combination of LLMs and XAI methods may be able to: (1) screen words associated with DIF during the item-writing process for immediate revision, (2) help review traditional DIF results by highlighting key words in the text, or (3) even replace the traditional DIF analysis process altogether if the prediction accuracy is sufficient.

This paper contains two investigations. The purpose of

## Figure 1

*Conceptual feedback loop to remove DIF during item writing. LLMs are used to detect DIF, then XAI methods are used to identify biased item text that needs revision. This process is completed without any field-test data.*

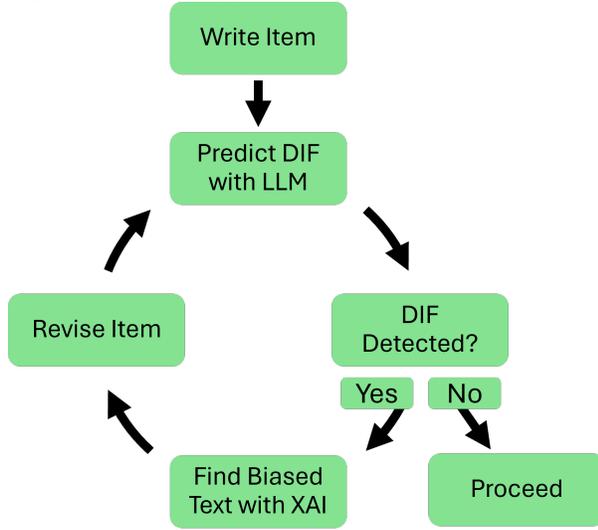

Study 1 is to compare and evaluate DIF prediction and XAI approaches to identify the methodology with the highest accuracy and interpretability of finding words associated with gender DIF from a quantitative perspective. Each method involves fine-tuning an encoder-based LLM to predict DIF between female and male examinees from the item text, and using XAI methods on the resulting model. After identifying the optimal approach, we apply it in Study 2 among eight DIF group pairs to examine the words that may influence DIF. Study 2 uses both a quantitative and qualitative evaluation.

### Background on DIF

Dichotomous and polytomous items in this paper were calibrated based on the two-parameter logistic model (Birnbaum, 1968) and generalized partial credit model (Muraki, 1992), respectively. The scores $\hat{\theta}$ based on these models were used to calculate the following DIF statistics for the respective item type, and used as labels for prediction.

### Dichotomous Items

For evaluating DIF among dichotomously scored items, the Mantel-Haenszel delta-difference (MH D-DIF) statistic (Dorans & Holland, 1992; Mantel & Haenszel, 1959) is commonly used. MH D-DIF for a particular item is expressed as

$$\text{MH D-DIF} = -2.35 \ln(\alpha), \tag{1}$$

which is a normalized transformation of the common odds ratio $\alpha$. To calculate $\alpha$, let $n_{Rsk}$ and $n_{Fsk}$ represent the number of reference and focal group examinees in the $s^{th}$ item score value in ability level group $k$. The reference group is often defined as the majority group which is suspected to have an advantage, while the focal group refers to the group suspected to have a disadvantage. Ability level groups are created based on deciles (i.e., ten equal size bins) of reference group $\hat{\theta}$. The focal group responses are categorized into those ability groups. For dichotomous items, $s \in \{0, 1\}$,

$$\alpha = \frac{\sum_k n_{R1k} n_{F0k}/n_{++k}}{\sum_k n_{F1k} n_{R0k}/n_{++k}}, \tag{2}$$

where the + indicates a summation over the particular index. For example, $n_{++k}$ is the number of all reference and focal examinees with any item score in ability group $k$. Standard error (SE) of MH D-DIF (Dorans & Holland, 1992; Holland & Thayer, 2013) is calculated as

$$\text{SE(MH D-DIF)} = 2.35 \sqrt{Var[ln(\hat{\theta}_{MH})]}, \tag{3}$$

and

$$Var[ln(\hat{\theta}_{MH})] = 2\left(\sum_k n_{R1k} n_{F0k}\right)^{-2} \sum_k (n_{R1k} n_{F0k} + \alpha n_{R0k} n_{F1k})$$
$$\times [n_{R1k} + n_{F0k} + \alpha(n_{R0k} + n_{F1k})]. \tag{4}$$

A negative, zero, and positive MH D-DIF shows that the item favors the reference group, no group, and focal group, respectively. The strength of DIF is interpreted as follows (Dorans & Holland, 1992; Zieky et al., 1993):

- Negligible DIF (A): |MH D-DIF| < 1 or not significantly different from zero

- Intermediate DIF (B): |MH D-DIF| ≥ 1, significantly different from zero, and |MH D-DIF| < 1.5 or not significantly greater than 1

- Large DIF (C): |MH D-DIF| ≥ 1.5 and significantly greater than 1 in absolute value

### Polytomous items

In this paper, for polytomous items scored in 3 or more categories, the effect size (ES) statistic based on the standardized mean difference (SMD) are used to evaluate DIF (Zwick & Thayer, 1996; Zwick et al., 1997):

$$\text{ES} = \frac{\text{SMD}}{\sigma_{\text{pooled}}} \tag{5}$$

and

$$\text{SMD} = \sum_k p_{Fk} m_{Fk} - \sum_k p_{Fk} m_{Rk}, \tag{6}$$

where $p_{Fk} = \frac{n_{F+k}}{n_{++k}}$ is the proportion of focal group members in group $k$, and $m_{Fk}$ and $m_{Rk}$ are the mean item scores for the respective groups. Ability level groups are formed based on $\hat{\theta}$ following the same procedure as the dichotomous items. The $\sigma_{\text{pooled}}$ is the pooled standard deviation (SD) of item scores, calculated as

$$\sigma_{\text{pooled}} = \sqrt{\frac{(N_F - 1)\sigma_F^2 + (N_R - 1)\sigma_R^2}{N_F + N_R - 2}}, \quad (7)$$

where $N$ is the number of examinees, and $\sigma_F^2$ and $\sigma_R^2$ are the group item score variance. The SE of SMD can be calculated based on the hypergeometric variance (Mantel, 1963):

$$\text{SE(SMD)} = \sqrt{\sum_k p_{Fk}^2 \left(\frac{1}{n_{F+k}} + \frac{1}{n_{R+k}}\right)^2 \text{Var}(F_k)}, \quad (8)$$

and

$$\text{Var}(F_k) = \frac{n_{R+k} n_{F+k}}{n_{++k}^2 (n_{++k} - 1)} \left\{ \left(n_{++k} \sum_s z_s^2 n_{+sk}\right) - \left(\sum_s z_s n_{+sk}\right)^2 \right\}, \quad (9)$$

where $z_s$ is the score value of score index $s$. For example, $\sum_s z_s n_{+sk}$ represents the sum of all scores in group $k$. The SE of SMD can be rescaled to the SE of ES using $\sigma_{\text{pooled}}$:

$$\text{SE(ES)} = \frac{\text{SE(SMD)}}{\sigma_{\text{pooled}}}. \quad (10)$$

A negative, zero, and positive ES shows that the item favors the reference group, no group, and focal group, respectively. The strength of DIF is interpreted as follows (Michaelides, 2008):

- Negligible DIF (A): |ES| < 0.17 or not significantly different from zero

- Intermediate DIF (B): 0.25 > |ES| ≥ 0.17 and significantly different from zero

- Large DIF (C): |ES| ≥ 0.25 and significantly different from zero

### Background on Transformer LLM

The introduction of Transformers (Vaswani et al., 2017) revolutionized natural language processing (NLP) by leveraging attention mechanisms to process entire text sequences in parallel, enabling faster training and scalability. BERT (Devlin et al., 2019) is the widely adopted Transformer-based LLM architecture, which advanced the field by offering pretrained models for downstream tasks such as text classification and question answering. Transformers excel at capturing contextual and semantic nuances, achieving state-of-the-art performance across benchmarks.

Despite the current popularity of decoder-only LLMs like GPT (T. B. Brown et al., 2020), encoder-only models like BERT have a significant efficiency advantage when the task is specialized and non-generative (Warner et al., 2024). Building upon BERT, DeBERTa (Decoding-enhanced BERT with disentangled attention) was introduced with innovations in attention mechanisms (He et al., 2021). Each word in DeBERTa is represented by two independent vectors: one for content and another for position. Using these vectors, the model computes the attention weights that take into account the interaction between word content and position. (He et al., 2021). For example, the words "artificial" and "intelligence" takes on a unique meaning when they occur adjacently, which is something DeBERTa can distinguish. In this paper, we model DIF by fine-tuning the pre-trained DeBERTa-v3-large model.

### Background on XAI

The goal of Explainable AI (XAI) is to enhance the interpretability of machine learning models, making them more transparent and trustworthy for human users (Hassija et al., 2024; Tjoa & Guan, 2021). As outlined by Tjoa and Guan (2021), XAI approaches typically aim to achieve one or more of the following objectives: (1) explaining the decisions or predictions made by a machine learning model, (2) finding the patterns within the internal mechanisms of the model, or (3) presenting the system with coherent models or mathematics.

In our study, XAI plays a crucial role in enhancing the utility and usability of a DIF prediction model by explaining the predictions made by Transformer LLMs and generating explanations that are mathematically coherent. There are two advantages in achieving these goals. First, it provides item writers with interpretable insights that can guide the creation of unbiased test items. By making the model's output understandable, item writers can interpret its predictions and adjust their items accordingly to ensure fairness. Second, it helps identify potential model misclassifications. By providing a clear understanding of the model's decision-making logic, it helps reduce the likelihood of item writers being misled by incorrect predictions, ensuring greater accountability of the model.

However, explaining the predictions of Transformer-based LLMs presents several unique challenges, particularly when dealing with textual data. For example, one broad area of XAI approaches explains the decision of a model by assigning values or creating an interpretable model that reflect the importance of features in their contribution to that decision, which includes the *saliency* methods (Tjoa & Guan, 2021), explanation by *simplification* (Angelov et al., 2021), or explanation by *feature relevance/importance* (Angelov et al., 2021). However, understanding the importance of specific words requires analyzing their role within the context of a sentence, as their significance is highly dependent on the surrounding text. More specifically, these models interpret tokens (i.e., word or part of a word) based on their text as well as their positions in the phrase.. This makes it inherently difficult to determine variable importance, as such words can appear in any position within a sentence.

To address these challenges, there are numerous post-hoc XAI techniques, both model-specific and model-agnostic, that can be applied to Transformer-based LLMs (Hassija et al., 2024; Tjoa & Guan, 2021; Zhao et al., 2024). The examples of such post-hoc XAI techniques include LIME (Locally-Interpretable Model-Agnostic Explanations; Ribeiro et al., 2016), SHAP (SHapley Additive exPlanations; Lundberg & Lee, 2017), and IG (Integrated Gradients; Sundararajan et al., 2017). LIME and SHAP are model-agnostic approaches which involve manipulating the original input data and examining how changes impact the prediction results, thereby attributing the magnitude and direction of each variable's effect on the model's predictions. On the other hand, IG is a model-specific approach to estimating the effect of each variable by integrating the gradients at all points along the path from the baseline to the original input.

An explanation needs to be mathematically sound yet easily understood. Thus, many XAI techniques are *additive feature attribution methods*, which use a simple explanation model that is a linear function of binary variables (Lundberg & Lee, 2017). Suppose $f$ is the original prediction model and $g$ is the explanation model which approximates $f$ using simplified inputs $z' \in \{0, 1\}^M$ of the original input vector $x$, where $M$ is the size of the original input vector. Then, the linear explanation model $g(z')$ is defined as:

$$g(z') = \phi_0 + \sum_{t=1}^{M} \phi_t z'_t, \tag{11}$$

where $\phi_t \in \mathbb{R}$ is the feature attribution value (also called SHAP value when using SHAP) for feature $t$. It is desirable that $g(z')$ exactly matches $f(x)$ when $z' = (1, 1, \ldots, 1)$, which is called the *local accuracy* property, and local methods try to approximate $f(x)$ whenever $z' \approx (1, 1, \ldots, 1)$. In addition to local accuracy, *missingness* and *consistency* are thought to be desirable properties by (Lundberg & Lee, 2017). Although all LIME, SHAP, and IG are additive feature attribution methods, only SHAP, which is based on the exact Shapley values, satisfies all the three properties. Furthermore, Lipovetsky and Conklin (2001) found that the Shapley values are very robust against multicollinearity in interpreting each feature's influence on the predicted variable. Therefore, in this paper, we use SHAP as our XAI approach. Being the default option of the `shap` package (Lundberg & Lee, 2017), the SHAP implementation we use is the Owen values (Owen, 1977) computed by using the `Explainer` class with the `algorithm = "partition"` option.

## Study 1: Comparison of Categorical and Continuous Models

The goal of Study 1 is to identify the methodology that offers the best combination of accuracy and interpretability of finding words associated with DIF. Using gender DIF data, we fine-tune models that predict DIF from the item text, and apply SHAP on the resulting model. Then we compare the utility of SHAP in identifying keywords associated with the DIF prediction models.

Although data were available for seven other DIF group pairings (see Study 2 for details), we focused exclusively on gender in Study 1 as it had the largest sample size in terms of both items and the number of focal and reference students per item. Therefore, we expected the gender data to provide the most reliable and generalizable results.

### Item Data

We used 42,176 items designed for English language arts and mathematics state summative assessments among students in grades 3 to 11. These items were written in accordance with the fairness Standards for Educational and Psychological Testing (AERA et al., 2014) by removing unnecessary barriers to the success of diverse groups of test takers. These basic guidelines were followed: (1) avoid measuring irrelevant constructs that advantages or disadvantages particular subgroups, (2) avoid angering, offending, upsetting, or distracting test takers, and (3) show respectful representation and treatment of all representatives of various cultures and backgrounds. Therefore, these items were designed to avoid bias and DIF. We expect any remaining bias to be rare and difficult to identify both objectively and subjectively.

All items have been previously field-tested and calibrated among grade-appropriate public school students in 20 states and territories in the United States from 2013 to 2023. Up to 10 unscored field-test items were embedded within a test with 20 to 35 scored items. Exams were administered in a computer adaptive testing format, but the field-test items were administered randomly (i.e., non-adaptive). Most students took both the English language arts and mathematics exams. About 30% of these items are no longer in operational use for various reasons, including but not limited to (1) failing to pass field-testing due to DIF or poor item statistics, or (2) they were publicly released.

To prepare the data for modeling, all item text was concatenated with a `[SEP]` separator between prompts and any available answer options. Only multiple-choice (33%), multiple-select (14%), and hot-text (8%) items had item options. Any reading or listening passages were excluded from the item text data due to their extensive lengths and the token limit of our models, which affected 37% of items. The maximum token size was set at 512, which caused 1% of items to have text removed from their tail ends ($M$ tokens=124,

$SD$=112).

MH D-DIF was used to quantify gender DIF for dichotomous items, and ES was used for polytomous items. To be able to model and interpret these two statistics nearly equivalently, ES (and its SE) was divided by 0.17. Therefore, for all items in the study, intermediate DIF thresholds of $\leq -1$ and $\geq 1$ were used and interpreted as favoring the reference group (male students; 2.7% of items) and focal group (female students; 3.3% of items), respectively. Large DIF thresholds of $\geq 1.5$ and $\leq -1.5$ were not considered because the data did not contain examples of these items (0.6% and 0.6% of items, respectively). Items having less than 100 examinees per focal and reference group per item were excluded from the analyses ($M$ female $n = 1,372$, $M$ male $n = 1,323$). $M(SD)$ gender DIF was 0.00(0.52).

### DIF Prediction

We partitioned the data randomly into approximately 80% training, 10% validation, and 10% test data. Items designed as a testlet with common passages were always assigned to the same data group. DeBERTa V3-large Transformer LLM (He et al., 2021) was fine-tuned to predict the DIF metric from the item text in Python using the `pytorch` library (Paszke et al., 2019) and a single NVIDIA A10G Tensor Core 24GB graphics processor (`batch size=8`, `learning rate=4e-6`, `weight decay=.001`). Training items were used to fine-tune the model, while the validation data was used to find the model with the lowest loss across epochs, which resulted in keeping the 2nd epoch for all four models described below. Test data were used to evaluate and report the prediction results.

#### Continuous Model

In the continuous model, we treat DIF as a continuous variable, and predict it from the item text. The mean-squared-error (MSE) loss is minimized in the fine-tuning process, which quantifies the distance between the target and predicted DIF values:

$$MSE = \frac{\sum (\hat{Y}_i - Y_i)^2}{N},\qquad(12)$$

where $Y_i$ is the target DIF statistic from the input data for item $i$, $\hat{Y}_i$ is its predicted value, and $N$ is the number of items.

#### Categorical Model

We also model DIF as a categorical variable in order to utilize its intermediate DIF cutoff values of $\geq 1$ and $\leq -1$. This ensures the model to be more accurate around these boundaries. Instead of predicting the categories directly, which would have resulted in severely unbalanced groups, we predict the probability of being classified into one of three DIF group categories (favoring reference, focal, or no group) upon repeated experiments. The probability of classifying item $i$ into DIF category $g$, denoted as $P_{ig}$, is derived based on the sampling distribution of the target DIF statistic, which is assumed to be normal with $\mu = Y_i$ and $\sigma^2 = SE_i^2$. The probability $P_{ig}$ is computed by using the cumulative distribution function, $\Phi_i$, for $\mathcal{N}(Y_i, SE_i^2)$:

$$P_{ig} = \begin{cases} \Phi_i(-1), & \text{if } g = \text{Reference group} \\ \Phi_i(1) - \Phi_i(-1), & \text{if } g = \text{No DIF group} \\ 1 - \Phi_i(1), & \text{if } g = \text{Focal group} \end{cases}\qquad(13)$$

where -1 and 1 represent the DIF group cutoff thresholds. Model output logits $L_{ig}$ were passed through a softmax function to calculate $\hat{P}_{ig}$, the predicted value of $P_{ig}$

$$\hat{P}_{ig} = \frac{e^{L_{ig}}}{\sum_g e^{L_{ig}}},\qquad(14)$$

so that $\sum_g \hat{P}_{ig} = 1$. The cross entropy loss (CEL) is minimized during the fine-tuning process, which quantifies the distance between target and predicted probabilities:

$$CEL = \frac{-\sum_i \sum_g P_{ig} \log \hat{P}_{ig}}{N}.\qquad(15)$$

From here on, item subscript $i$ is suppressed from all expressions for succinctness (e.g., $P_{ig} \equiv P_g$).

#### Alternative Random Seed Models

Fine-tuning Transformer LLMs involves stochastic optimization. We suspected that this could heavily influence the outcomes of SHAP, as XAI methods are often sensitive to minor variations and may produce inconsistent results (Pirie et al., 2023). To investigate this, we replicated the continuous and categorical models with an additional different random seed prior to fine-tuning. Therefore, a total of four models were fine-tuned.

### XAI for the Continuous Model

The `Explainer` class from the `shap` python package (Lundberg & Lee, 2017) was used to calculate token attributions for all models (`fixed_context=1`, `algorithm="partition"`). For each item $i$, we calculated attribution values $\phi_t$ for every token $t$ to the predicted DIF $\hat{Y}$. Each item can have up to 512 tokens. Token attributions are similar to regression slopes, and is on the same scale as $\hat{Y}$. For example, an attribution value of 0.1 means that the token contributes to the prediction $\hat{Y}$ by 0.1.

### XAI for the Categorical Model

Unlike the continuous model, the categorical model outputs three probabilities per item, representing the three DIF

groups. Therefore, we also output three attribution values per token, represented as $\phi_{tg}$. For example, given the current array of the tokens, a $\phi_{tg} = 0.01$ indicates that the inclusion of a token $t$ for the item is expected to increase the probability of being assigned to group $g$ by 1%, while negative values indicate increased chance of belonging in the other two groups. Given these characteristics, we found considerable redundancy in the three attribution values. The correlations between them were $r_{(\phi_{t1}, \phi_{t2})} = -.57$, $r_{(\phi_{t1}, \phi_{t3})} = -.05$, and $r_{(\phi_{t2}, \phi_{t3})} = -.79$. In order to simplify their interpretation, we convert these to a single value per token $\phi_t$ by

$$\phi_t = \begin{cases} -\phi_{t1} & \text{if } \phi_{t1} > 0, \\ 0 & \text{otherwise} \end{cases} + \begin{cases} \phi_{t3} & \text{if } \phi_{t3} > 0, \\ 0 & \text{otherwise,} \end{cases} \quad (16)$$

where $\phi_{t1}$ is the token attribution for the reference group, and $\phi_{t3}$ is the focal group. As the equation shows, we ignored $\phi_{t2}$ (i.e., no DIF group) because it tended to capture the noise in the DIF data, and focusing on the other two groups isolated and retained the important information. We also ignored negative values for $\phi_{t1}$ and $\phi_{t3}$ as they indicated an association with the other two groups, which was redundant. As a result, in both the continuous and categorical models, a negative $\phi_t$ indicated favoring the reference group (male students), while a positive $\phi_t$ indicated favoring the focal group (female students). Correlations of attributions with the original three attributions were $r_{(\phi_t, \phi_{t1})} = -.46$, $r_{(\phi_t, \phi_{t2})} = -.34$, and $r_{(\phi_t, \phi_{t3})} = .75$.

### Evaluation of Prediction and XAI Methods

Using the test data, accuracy, simplicity, and interpretability of the prediction and attribution values were compared between models. Output predicted values were entered into a multiple regression model to re-predict $Y$ to calculate $R^2$. For the continuous models, $\hat{Y}$ was the sole explanatory variable. For the categorical models, the logit of the three probability values $\hat{P}_g$ were entered as explanatory variables, including all interaction terms. This is a simple approach to compare all models with $R^2$, especially because CEL is difficult to interpret. Higher $R^2$ values indicate a better prediction, which also suggests that the attribution values are more accurate.

There are no agreed upon methods to evaluate post-hoc XAI metrics, but they should quantify the simplicity and accuracy of the interpretations (Zhou et al., 2021). Attributions were evaluated in the following manner. First, token attributions should be unbiased, such that when $Y = 0$, average attribution should also be 0 when the baseline predicted probabilities (i.e., $\phi_0$) is 0. Therefore, we calculated bias by taking the mean of $\phi_t$ for 100 items with $Y$ values closest to 0. Further, given the properties of the attribution values $\phi_t$, it should be positively correlated with $Y$. Additionally, a peaky marginal distribution of $\phi_t$ is desirable, where most tokens have near-zero influence on the predicted DIF, while very few tokens have extreme attribution values and high influence. In practice, this would simplify not only the interpretation, but also the item revision as only a few words will need to be replaced to eliminate DIF. This pattern can be represented with a high kurtosis. Although SD is another option, we prefer kurtosis as SD depends on the scale of both $\phi_t$ and the target variable ($Y$ or $P_g$), while kurtosis is standardized. Therefore, kurtosis may be easier to interpret than SD when comparing models. Finally, to examine the stability of $\phi_t$, attribution values from the seed replications were evaluated using correlations (which represent reliability $\rho_{\phi_t}$) and root-mean-squared-error (RMSE).

We also tested whether the attribution values can accurately be interpreted like a regression coefficient by replacing an influential token with [UNK], which is the special token for an unknown word. In theory, the predicted values should decrease by approximately the equal amount as the replaced token attribution value. For a categorical model from a single random seed, the lowest attribution token (i.e., favoring males) was replaced for each item, and the change in $\hat{P}_1$ (i.e., original minus new $\hat{P}_1$) was compared with the original attribution value using correlation, RMSE, and bias. Similarly, the highest attribution token (i.e., favoring females) was replaced, and the change in $\hat{P}_3$ was compared with the original attribution value. Further, for a continuous model from a single random seed, The highest and lowest attribution tokens were replaced separately, and the change in $\hat{Y}$ was compared to the attribution value.

### Study 1: Results

The $R^2$ for the test data was .33, .33, .31, and .31, for the two continuous models and the two categorical models, respectively, showing similar predictive power across all four models. The $R^2$ for the average of the two seed alternatives for the continuous and categorical models showed no meaningful change (.33 and .32, respectively). Attribution value bias was minimal for all models (between -0.0001 and 0.0002; see Table 1). The correlation of attribution values between two random seed alternative models was $r = .66$ (RMSE=0.016) for continuous and $r = .60$ for categorical models (RMSE=0.0031). Based on these correlations and the Spearman-Brown Prophecy formula (W. Brown, 1910; Spearman, 1910), the reliability of the average of two models was $\rho_{\phi_t} = .80$ and .75 for continuous and categorical models, respectively. The improvement in reliability resulted in about a 10% increase in $r_{(\phi_t, Y)}$ for both models. This shows that fine-tuning and averaging multiple models can improve the accuracy of the attribution values. Correlations between the attributions and DIF showed that interpreting individual attributions from the categorical model ($r = .200$) is more accurate and less noisy compared to the continuous model ($r = .088$). Kurtosis was also higher for the categorical model (515) than continuous (90), showing easier interpre-

**Table 1**

*Comparing Gender Token Attributions between Models*

| Model | Seed | M | SD | Kurtosis | Bias | $r_{(\phi_t, Y)}$ |
|---|---|---|---|---|---|---|
| Categorical | 1 | .000 | .004 | 736 | -0.0001 | .178 |
| | 2 | .000 | .003 | 399 | -0.0001 | .183 |
| | Averaged | .000 | .003 | 515 | -0.0001 | .200 |
| Continuous | 1 | .000 | .020 | 91 | 0.0002 | .080 |
| | 2 | .000 | .018 | 83 | -0.0001 | .081 |
| | Averaged | .000 | .017 | 90 | 0.0001 | .088 |

*Note.* Bias was calculated by taking the mean attribution for 100 test data items with DIF values closest to 0.

tation of the categorical model as fewer tokens had extreme attribution values.

Token replacement results show that the attribution values from the categorical model can better predict the change in the predicted values when the most influential tokens are replaced (see Table 2). Correlation values were stronger for the categorical model compared to the continuous model. These make the attribution values from the categorical model more straight forward to interpret than the continuous model.

**Table 2**

*Token Attribution Accuracy Evaluated by Replacing the Lowest and Highest Attribution Tokens for Every Item*

| Model | Replaced Token | $r$ | RMSE | Bias |
|---|---|---|---|---|
| Categorical | Favoring male | -.71 | 0.026 | 0.010 |
| | Favoring female | .95 | 0.011 | 0.003 |
| Continuous | Favoring male | .28 | 0.102 | 0.063 |
| | Favoring female | .90 | 0.053 | 0.028 |

*Note.* RMSE = root-mean-squared error. Correlation, RMSE, and bias compares the replaced attribution value and the change in the predicted value (i.e., original minus new prediction). Correlations are expected to be positive, except when the token favoring males was replaced for the categorical model. RMSE and bias are not directly comparable between the models as they are on different scales. Bias was calculated with by taking the mean of the prediction change minus the replaced attribution.

In summary, we found that the categorical model was superior to the continuous model in its accuracy, interpretability, and simplicity. The inherent inclusion of the prediction uncertainty in the categorical model (e.g., classification probability) is valuable as well, which is missing from the continuous model. Also, two models fine-tuned with a different random seed led to unstable results, but averaging these two models improved the reliability considerably. We call the combination of two categorical models the "averaged categorical model". The averaged categorical model was the optimal DIF prediction and XAI approach we identified in

Study 1.

## Study 2: Application of the Averaged Categorical Model to Eight DIF Groups

In this section, we apply the averaged categorical model to eight DIF group pairs separately. Other than the inclusion of additional groups, the item data and general procedures were identical to the averaged categorical model approach in Study 1. The focal/reference group pairs included (1) female/male, (2) Asian and Pacific Islanders (Asian)/White, (3) Black or African American (Black)/White, (4) Hispanic or Latino (Hispanic)/White, (5) Native American and Alaskan Native (Native)/White, (6) lower social economic status (LSES)/Non-LSES, (7) students with disabilities (SWD)/Non-SWD, and (8) English learners (EL)/Non-EL. Students associating with multiple races were excluded from race-based DIF analyses because they could not be classified into one group. Items having less than 100 examinees per focal and reference group were excluded from the analyses for the respective group pair. A total of 42,180 items had at least one DIF group data available. However, the number of students per item for the focal group was sometimes very low, which limited the number of included items and increased the DIF SE (see Table 3). In particular, Native American and Alaskan Native group had the lowest number of items (4,682) and students per item (166).

Identical to Study 1, data were randomly partitioned into approximately 80% training, 10% validation, and 10% test data. After fine-tuning, SHAP procedure was applied and used to find prominent words that were associated with DIF. To find the top 10 tokens favoring the focal group, tokens with $\phi_t < 0.01$ and non-alphabet characters were removed from the test data. Tokens were then converted to all lower case. Tokens occurring less than 3 times were also removed. Finally, the top tokens were found based on the highest within-token mean of $\phi_t$ across all items. Similarly, the top tokens favoring the reference group were found by removing the data with $\phi_t > -0.01$, and finding the tokens with the lowest mean of $\phi_t$.

**Table 3**

*Descriptive Statistics*

| Foc/Ref Group | $N$ Items | $N$ Students[a] | | DIF | | $P_g$[a] | | |
| | | Foc | Ref | $M(SD)$ | $SE$[a] | Foc | No DIF | Ref |
|---|---|---|---|---|---|---|---|---|
| Female/Male | 42,176 | 1,323 | 1,372 | 0.00(0.52) | 0.27 | .05 | .91 | .04 |
| Asian/White | 34,515 | 253 | 1,201 | 0.13(0.72) | 0.46 | .14 | .77 | .08 |
| Black/White | 32,617 | 244 | 1,271 | -0.10(0.59) | 0.46 | .07 | .83 | .11 |
| Hispanic/White | 41,695 | 1,014 | 1,087 | -0.12(0.49) | 0.31 | .03 | .91 | .06 |
| Native/White | 4,682 | 166 | 2,745 | -0.07(0.40) | 0.47 | .04 | .89 | .07 |
| LSES/Non-LSES | 42,175 | 1,404 | 1,292 | -0.13(0.39) | 0.27 | .01 | .95 | .04 |
| SWD/Non-SWD | 35,049 | 324 | 2,731 | -0.16(0.56) | 0.40 | .04 | .87 | .09 |
| EL/Non-EL | 27,089 | 438 | 2,810 | -0.30(0.89) | 0.47 | .07 | .75 | .18 |

*Note.* DIF = differential item functioning, Foc = focal group, Ref = reference group, Asian = Asian and Pacific Islanders, Black = Black or African American, Hispanic = Hispanic or Latino, Native = Native American and Alaskan Native, LSES = lower social economic status, SWD = students with disabilities, EL = English learners. [a]Averaged across items.

### Study 2: Results

Results from the test data are reported in this section. The $R^2$ varied considerably depending on the group (see Table 4). For example, the female/male group simultaneously had the highest (1) number of items, (2) sample size per item, and (3) $R^2$ of .32. On the contrary, all three of these statistics were the lowest among the Native/White group. One reason for these disparities is the small sample size, which leads to poor generalizability, affecting model performance on test data. An alternative reason is that DIF items were more common among some groups than others, as evidenced by the varying DIF $SD$ ranging from 0.39 to 0.89. DIF with a higher $SD$ is presumably easier to predict because there are more items with extreme DIF effects. The final possible reason for the group differences in $R^2$ is that DIF for some groups are simply more difficult to predict based on the text. For example, Hispanic/White group had similar sample sizes and DIF $SD$ as female/male but had much lower $R^2$ of .16. Additionally, differences in the prediction RMSE and correlations showed that DIF favoring the reference group is sometimes easier to predict than the focal, or vice versa. For example, items favoring female students were easier to predict (RMSE=0.11, $r$=.62) than male students (RMSE = 0.14, $r$=.33).

Mean attribution values were near 0 for all groups (range from -.001 to .000; see Table 5). The poor overall model fit of the Native/White group was reflected in the low $r_{(\phi_i, Y)}$ = .02 and $\rho_{\phi_i}$ = .21. Kurtosis was highest for SWD/Non-SWD, and $r_{(\phi_i, Y)}$ was highest for the female/male group. Other than the Native/White group, $\rho_{\phi_i}$ was near .70 for all (range from .61 to .75).

### Qualitative Results

Example item text along with their token attribution values are plotted in Figure 2. These items all had observed DIF of < -1 or > 1, which may warrant item review in traditional DIF analyses. Highlighting text based on token attributions may assist in understanding what may have caused the DIF flagging to occur. For example, the fourth item on the figure shows "dog walks" as a phrase that favors non-SWD students, potentially because they are able to walk dogs, while some SWD students may have walking limitations. However, many of the token attributions are still difficult to interpret qualitatively. Nevertheless, visualizations in this format may be helpful for item developers for pinpointing, understanding, or revising the potential sources of DIF.

Top 10 tokens favoring with each focal and reference groups are listed in Table 6, although some groups had less than 10 tokens that qualified. Some words simply seemed spurious and not meaningful (e.g., a, two, for, is). The literature recognizes that Transformer models attend to irrelevant information (Ye et al., 2024), which could explain these attribution values that seem like noise. Interestingly, names such as "John", "Karen", and "Aaron" tended to favor White students. However, many of these influential words were directly related to the English language arts subject (e.g., text, summarize) or mathematics (e.g., multiplication, equation), suggesting that the DIF associated with these words may be minor sub-domains included in the test blueprint by design, rather than construct-irrelevant item content that should be removed from assessments.

### Discussion

To the authors' knowledge, this is the first paper to predict DIF using LLMs or NLP, and also the first to identify words associated with DIF using XAI. DIF prediction accuracy was low to moderately high depending on the group. Our method may help improve item development by providing a type of contextual feedback that was previously unavailable. Our

**Figure 2**

*Four example items with DIF. Words are highlighted based on averaged categorical model token attribution values.*

Observed DIF=1.06
Predicted favoring:
Asian p=0.30
White p=0.02

Select the two sentences that are punctuated correctly.[SEP]While I was growing up in the Midwest my favorite question to hear from my parents was "Guess where we're going this time?"[SEP]Although by that point, my parents had the whole vacation planned out; the moment they told me, I started looking up the location to see what activities were available.[SEP]When I was eight my family voted on a vacation to New York City where we stayed in downtown Times Square. Then later when I was ten we flew to Florida again, this time we departed on a cruise to Mexico, Jamaica and the Bahamas for a second time.[SEP]The average life expectancy is seventy years on this planet, this planet has so many different geological features, different climates and different cultures.[SEP]The places I have already visited make my curiosity even greater, and I think that it's important to view the world and ways of life from a different point of view.[SEP]Last year when I was sixteen we went on another cruise where we sailed the Western Caribbean to Puerto Rico, the Bahamas yet again and St Thomas.

Observed DIF=1.04
Predicted favoring:
Female p=0.71
Male p=0.00

What inference can be made about the narrator's feelings toward the new traveling companions? Support your answer with details from the text.

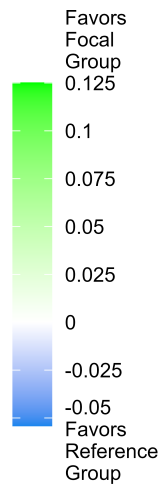

Favors
Focal
Group
0.125
0.1
0.075
0.05
0.025
0
-0.025
-0.05
Favors
Reference
Group

Observed DIF=-1.04
Predicted favoring:
Hispanic p=0.02
White p=0.19

A student is writing a story for his English class about a person who shows dedication. Read the draft of the story and answer the question that follows.When Luke was a little boy, he sat on the sidelines of the high school football games and watched his uncle coach the team on to victory. That was a weekly tradition for years; every Friday night at the local game, he could be found on the sidelines. During his elementary school days, winning came easily to the high school team. As Luke entered middle school, the team began to lose more frequently, but their winning spirit never died.Now Luke was in high school. No longer cheering from the sidelines, he would be on the field. The community had drastically changed. Big employers had moved out of town, and the high school enrollment dropped from 1,200 to 400. The availability of tough football players had also decreased, but Luke's dedication to the team was unchanged. Every week he ran proudly onto the football field. He played every game of his high school career, but sadly, the team had a losing record. In his senior year, Luke was the team captain and worked hard to condition the team and himself. Every point scored was celebrated as though it were a world championship game. The "win" was his dedication to the team, not the number of points on the scoreboard.What is a more specific word to replace the underlined word?[SEP]suffered[SEP]obtained[SEP]received[SEP]permitted

Observed DIF=-1.78
Predicted favoring:
SWD p=0.03
Non-SWD p=0.39

What is the median number of dog walks for the first 9 weeks?

**Table 4**

*Prediction Summary for the Averaged Categorical Model*

| Foc/Ref Group | CEL | $R^2$ | $\hat{P}_1$ (Favors Reference) | | | | $\hat{P}_3$ (Favors Focal) | | | |
|---|---|---|---|---|---|---|---|---|---|---|
| | | | $M(SD)$ | $r$ | RMSE | Bias | $M(SD)$ | $r$ | RMSE | Bias |
| Female/Male | 0.12 | .32 | .04(.05) | .33 | 0.14 | -0.01 | .04(.09) | .62 | 0.11 | 0.00 |
| Asian/White | 0.22 | .20 | .07(.04) | .26 | 0.18 | -0.02 | .13(.10) | .47 | 0.21 | -0.01 |
| Black/White | 0.19 | .11 | .10(.05) | .27 | 0.18 | -0.01 | .05(.03) | .28 | 0.13 | -0.01 |
| Hispanic/White | 0.12 | .16 | .05(.05) | .32 | 0.15 | -0.01 | .02(.02) | .23 | 0.10 | -0.01 |
| Native/White | 0.13 | .04 | .06(.02) | .14 | 0.10 | 0.00 | .04(.02) | .22 | 0.09 | -0.01 |
| LSES/Non-LSES | 0.07 | .12 | .04(.04) | .30 | 0.11 | 0.00 | .01(.01) | .15 | 0.05 | 0.00 |
| SWD/Non-SWD | 0.15 | .11 | .08(.07) | .39 | 0.17 | 0.00 | .03(.02) | .22 | 0.10 | -0.01 |
| EL/Non-EL | 0.22 | .08 | .17(.10) | .28 | 0.29 | -0.01 | .06(.03) | .20 | 0.17 | 0.00 |

*Note.* Foc = focal group, Ref = reference group, CEL = cross-entropy loss, DIF = differential item functioning, Asian = Asian and Pacific Islanders, Black = Black or African American, Hispanic = Hispanic or Latino, Native = Native American and Alaskan Native, LSES = lower social economic status, SWD = students with disabilities, EL = English learners.

**Table 5**

*Token Attribution Summary for the Averaged Categorical Model*

| Foc/Ref Group | $M$ | $SD$ | Kurtosis | $r_{(\phi_t, Y)}$ | $\rho_{\phi_t}$ |
|---|---|---|---|---|---|
| Female/Male | .000 | .003 | 515 | .20 | .75 |
| Asian/White | .000 | .004 | 793 | .12 | .75 |
| Black/White | -.001 | .002 | 316 | .08 | .68 |
| Hispanic/White | -.001 | .002 | 288 | .14 | .70 |
| Native/White | .000 | .001 | 453 | .02 | .21 |
| LSES/Non-LSES | .000 | .001 | 213 | .11 | .70 |
| SWD/Non-SWD | -.001 | .003 | 1329 | .09 | .61 |
| EL/Non-EL | -.001 | .004 | 177 | .07 | .67 |

*Note.* Foc = focal group, Ref = reference group, Asian = Asian and Pacific Islanders, Black = Black or African American, Hispanic = Hispanic or Latino, Native = Native American and Alaskan Native, LSES = lower social economic status, SWD = students with disabilities, EL = English learners. $\rho_{\phi_t}$ is the reliability of token attributions between two models with different random seeds during fine-tuning.

approach can be used to (1) screen words associated with DIF during the item-writing process for immediate revision, (2) help review traditional DIF analysis results by highlighting key words in the text, or (3) even replace the traditional DIF analysis process altogether if the prediction accuracy can be raised sufficiently high in the future. Extensions of this research could help further the fairness of assessment programs, especially those that lack resources to build high-quality items, and among smaller demographic groups where reaching sufficient sample sizes for traditional DIF analyses is difficult.

One word of caution when interpreting attribution values is that they represent a correlational rather than a causal relationship between the word and DIF. For example, even if the phrase "read the following story" had a high attribution value, removing these four words may not result in reduced observed DIF. Instead, the DIF may be due to one group excelling at reading stories, in which case the entire story may be the cause of DIF. The association between the word and DIF may be ambiguous and subjective in many cases.

In our study, most words associated with DIF seemed to reflect construct-relevant minor multidimensionality inherent in the test blueprint rather than construct-irrelevant item bias that warrants removal from assessments (Osterlind & Everson, 2009). For example, words that favored female students tended to appear on writing items (e.g., summarize). This may simply indicate that female students tend to be better writers than male students, given the same overall test score, which has been observed in other studies (e.g., Zhang et al., 2019). If this is true, male students are likely performing better at the non-writing items than female students, but only the writing items are flagged for DIF because they make up a fewer number of items in the exam. This could explain why qualitative reviews of DIF items often yield confusing or inconclusive results. Addressing the effects of minor multidimensionality could remove a major source of noise from DIF analysis that make unwanted bias difficult to identify (Bulut & Suh, 2017; Camilli, 1992).

The use of AI has been gaining increased traction in the broader context of assessment development (Hao et al., 2024; Jiao et al., 2023). For example, AI has been applied to item generation (Gierl et al., 2012), field-testing (Maeda, 2024), and scoring (Lottridge et al., 2023). However, AI has simultaneously become a growing concern (Burstein et al., 2024). AI and LLMs are commonly recognized as being biased (Bolukbasi et al., 2016; Mehrabi et al., 2021). This

**Table 6**

*Top 10 Ranking Tokens Favoring Focal and Reference Groups*

| Foc/Ref Group | Tokens favoring the focal group | Tokens favoring the reference group |
| --- | --- | --- |
| Female/Male | narrator, message, text, reader, summarize, relationship, inference, intend, feelings, evidence | growth, decay, equal, option, rounded, number, is, mean, hour, nearest |
| Asian/White | spelling, spelled, factor, capitalization, multiplication, punctuated, evidence, mistakes, errors, students | when, read, two, an, mr, click, to, sentence, consider, karen |
| Black/White | div, multiplying, quotient, multiplication, equation, divide, decimal, times, unknown, fraction | grams, aaron, equal, parts, an, enter, clock, kilograms, john, layers |
| Hispanic/White | enter, div, equation, quotient, rational, multiplication, expression, sum, exact, product | rounded, punctuated, shade, phrases, parts, growth, hundred, underlined, word, minute |
| Native/White | | a |
| LSES/Non-LSES | box, irrational, select | equal, rounded, measure, degrees, word, answer, use, number, performance, box |
| SWD/Non-SWD | div, size, unknown, equation, makes, true, for | directions, argumentative, farm, student, performance, club, s, covering, narrative, re |
| EL/Non-EL | from, equations | round, scored, task, mean, read, performance, sentence, word, select, meaning |

*Note.* Foc = focal group, Ref = reference group, Asian = Asian and Pacific Islanders, Black = Black or African American, Hispanic = Hispanic or Latino, Native = Native American and Alaskan Native, LSES = lower social economic status, SWD = students with disabilities, EL = English learners. Some groups had less than 10 qualifying tokens.

might be a reflection of the biases present in the training data, where AI learns to mimic human prejudices (Caliskan et al., 2017). Nevertheless, AI can be debiased or designed to detect bias when specifically and carefully trained for that purpose (Cheng et al., 2021; Garimella et al., 2021; Lauscher et al., 2021; Liang et al., 2020). Therefore, our AI-driven DIF detection approach may be a valuable and feasible framework for mitigating bias and promoting fairness in AI-based assessments. Note that our models were trained using human-written items that underwent rigorous quality control based on strict bias and sensitivity guidelines. As a result, these models may not be equally effective at detecting bias in AI-generated items, including blatant biases that are obvious to human reviewers.

**Limitations**

One of the challenges of training our own LLM is computational capacity. Fine-tuning a single LLM and calculating its SHAP attribution values can take multiple days. This was the main reason why models from only two different seeds were trained. Fine-tuning and averaging more than two models will further improve attribution value accuracy as a result of increased fine-tuning reliability, and is recommended if resources allow.

However, fine-tuning instability may go beyond the random seed we considered in this paper. For example, fine-tuning using slightly different hyperparameters or training data may have considerable influence on the attribution val-ues. Also, the original DIF target variable is already an estimate associated with error variance. In particular, fine-tuning a well-fitting model is extremely difficult for low-sample size groups with high DIF standard error, such as the Native American and Alaskan Native group.

Further, formatting item text for LLM input is often challenging. In our case, the item text was stored in HTML format within JSON data, which we had to convert to plain text before LLM consumption. Items included features like bold, underlined, or italicized text, as well as images, graphs, tables, audio, and various accessibility options. These elements that extend beyond plain text, long reading passages, and the correct answer keys had to be excluded from the input data. Additionally, when students take the exam on a computer, they interact with information positioned in specific screen locations (e.g., spacing of paragraphs and response options), which can influence their experience of the item. These factors make it difficult for AI to replicate the way human test-takers perceive and engage with items.

**Future Directions**

SHAP (Lundberg & Lee, 2017) is just one of the many XAI methods that exist in the literature. Testing others may be worthwhile as they tend to provide varied results (Pirie et al., 2023). Calculating and displaying standard error-like metrics to characterize the precision of token attributions may further help their interpretability among the lay audience (e.g., item writers). Our proposed approach may be further enhanced by including covariates in the prediction, which are variables linked to item difficulty but not captured by LLMs. These may include readability indices, text length (AlKhuzaey et al., 2023), or response time (Duan & Cheng, 2024).

A recently introduced encoder Transformer LLM called ModernBERT could address multiple limitations we faced in this paper (Warner et al., 2024). For example, its alternating attention, unpadding, and flash attention mechanisms provide considerable computational efficiency improvements. Having a 8,192 token limit, it can process items with very long reading passages. Lastly, ModernBERT was trained on web documents and code, so it may be able to understand raw HTML item text without the loss of nuanced item formatting information.

Similarly, a new LLM architecture called the Diff Transformer (Ye et al., 2024) may also be useful for furthering our research. Their paper recognizes that traditional Transformers tend to over-allocate attention to irrelevant information. This aligns with our observations in the current paper, especially with the continuous model. Diff Transformers are able to amplify model attention to the relevant context, while canceling irrelevant noise. This may translate to not only increased predictive power, but also improved precision of XAI methods in identifying words linked to DIF.

AI researchers continue to propose new methods that could advance our contributions in this paper. We remain eager and vigilant for these developments.